\documentclass[10pt,letterpaper]{article}

\usepackage[top=0.85in,left=2.75in,footskip=0.75in]{geometry}

\usepackage{amsmath,amssymb}

\usepackage{changepage}

\usepackage[utf8x]{inputenc}

\usepackage{textcomp,marvosym}

\usepackage{cite}
																	
\usepackage{nameref, hyperref}

\usepackage[right]{lineno}

\usepackage{microtype}
\DisableLigatures[f]{encoding = *, family = * }

\usepackage[table]{xcolor}

\usepackage{array}

\newcolumntype{+}{!{\vrule width 2pt}}

\newlength\savedwidth

\newcommand\thickhline{\noalign{\global\savedwidth\arrayrulewidth\global\arrayrulewidth 2pt}%
\hline
\noalign{\global\arrayrulewidth\savedwidth}}

\raggedright
\usepackage[aboveskip=1pt,labelfont=bf,labelsep=period,justification=raggedright,singlelinecheck=off]{caption}

\makeatletter
\renewcommand{\@biblabel}[1]{\quad#1.}
\makeatother

\usepackage{lastpage,fancyhdr,graphicx}
\usepackage{epstopdf}
\usepackage{comment}
\fancyheadoffset[L]{2.25in}
\fancyfootoffset[L]{2.25in}
\begin{document}
\vspace*{0.2in}

\begin{flushleft}
{\Large
\textbf\newline{A Comparative Study on Polyp Classification using Convolutional Neural Networks} 
}
\newline
\\
Krushi Patel\textsuperscript{1},
Kaidong Li\textsuperscript{1},
Ke Tao\textsuperscript{2},
Quan Wang\textsuperscript{2},
Ajay Bansal\textsuperscript{3},
Amit Rastogi\textsuperscript{3},
Guanghui Wang\textsuperscript{1\dag}
\\
\bigskip
\text{1} School of Engineering, University of Kansas, Lawrence, KS 66045
\\
\text{2} The First Hospital of Jilin University Changchun 130000, China
\\
\text{3} The University of Kansas Medical Center, Kansas City, KS 66160
\\

%
%
\dag Corresponding author. ghwang@ku.edu





\end{flushleft}

\section*{Abstract}
Colorectal cancer is the third most common cancer diagnosed in both men and women in the United States. Most colorectal cancers start as a growth on the inner lining of the colon or rectum, called `polyp’. Not all polyps are cancerous, but some can develop into cancer. Early detection and recognition of the type of polyps is critical to prevent cancer and change outcomes. However, visual classification of polyps is challenging due to varying illumination conditions of endoscopy, variant texture, appearance, and overlapping morphology between polyps. More importantly, evaluation of polyp patterns by gastroenterologists is subjective leading to a poor agreement among observers. Deep convolutional neural networks have proven very successful in object classification across various object categories. In this work, we compare the performance of the state-of-the-art general object classification models for polyp classification. We trained a total of six CNN models end-to-end using a dataset of 157 video sequences composed of two types of polyps: hyperplastic and adenomatous. Our results demonstrate that the state-of-the-art CNN models can successfully classify polyps with an accuracy comparable or better than reported among gastroenterologists. The results of this study can guide future research in polyp classification.


\section*{Introduction}
Colorectal cancer is the third most common cancer diagnosed in both men and women in the united states \cite{state}. According to the American Cancer Society, a total of 101,420 new cases of colon cancer and 44,180 new cases of rectal cancer occurred in 2019. The lifetime risk of developing colorectal cancer is about 4.99\% for men and 4.15\% for women \cite{state}. Colorectal cancer is the second leading cause of cancer-related deaths. Colon cancer is expected to cause about 51,020 death in the United States during 2020. 

Polyps are considered the harbinger of colorectal cancer. Early detection and recognition of polyps can reduce death caused by colorectal cancers. Broadly speaking, colorectal polyps can be divided into two categories: non-neoplastic (Hyperplastic) and neoplastic (Adenomatous) \cite{shinya1979morphology}.
Hyperplastic polyps do not predispose to cancer, whereas adenomatous polyps are considered pre-cancerous as they account for approximately 85\% \cite{kim} of sporadic colorectal cancers via the adenoma-carcinoma pathway. Therefore, adenomatous polyps are removed during colonoscopy to prevent future cancer. Therefore, differentiating the two types of polyp histology is critical to determine which patient needs close follow up at shorter intervals and which patient can be surveyed every 10 years.

\begin{figure*}[!h]
\begin{center}
\begin{tabular}{ccc}
 \includegraphics[width=30mm,height=26mm]{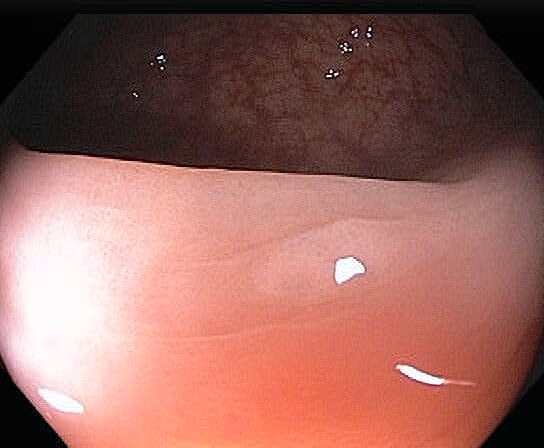} &  \includegraphics[width=30mm,height=26mm]{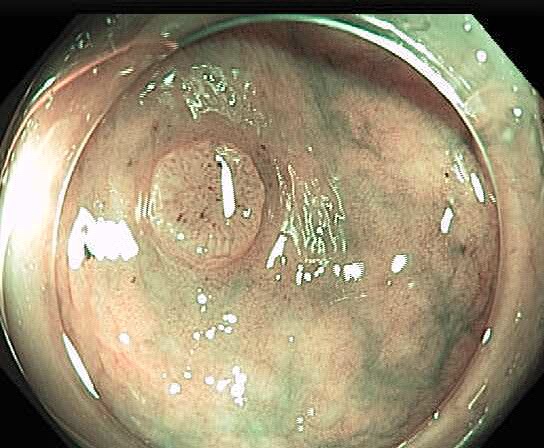} &  \includegraphics[width=30mm,height=26mm]{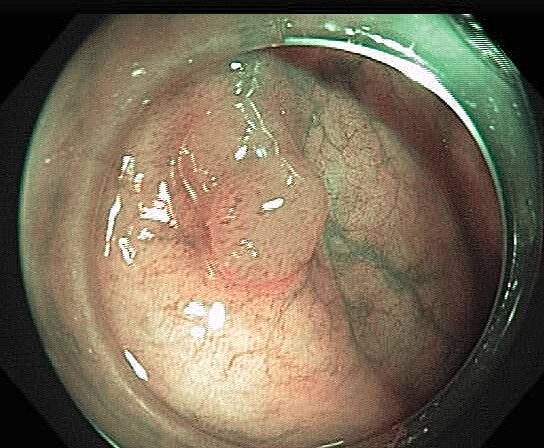} \\
 \includegraphics[width=30mm,height=26mm]{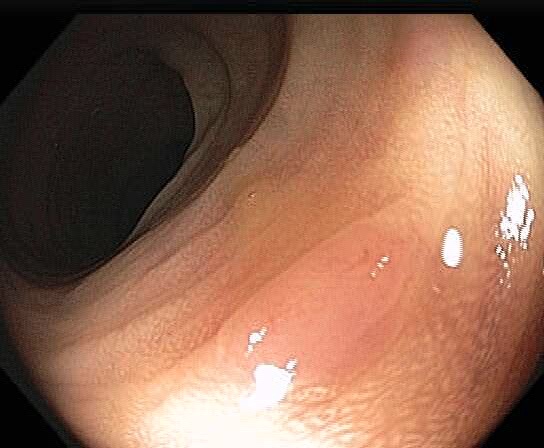} & \includegraphics[width=30mm,height=26mm]{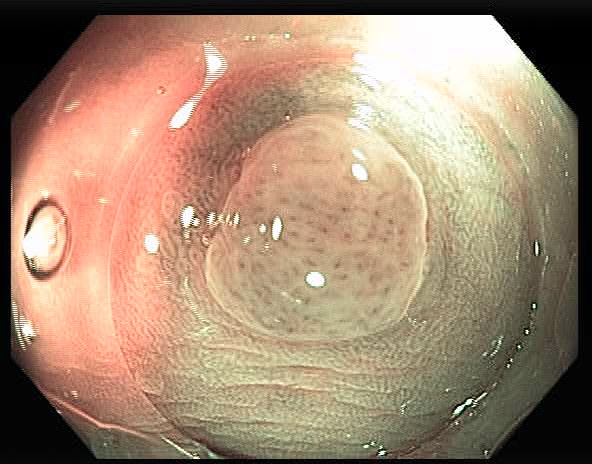} &   \includegraphics[width=30mm,height=26mm]{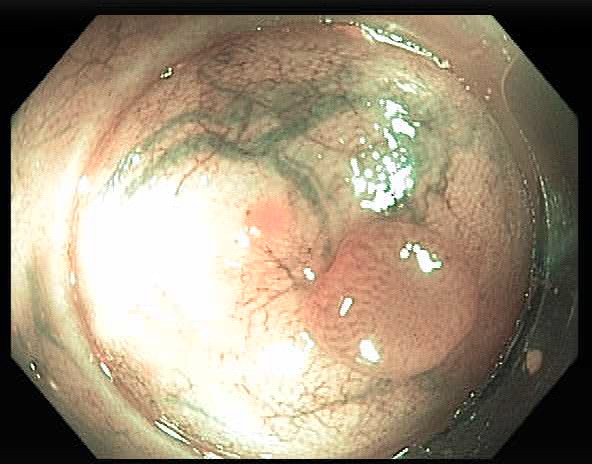} \\
\end{tabular}
\caption{{\bf Example of polyps from different class with subtle difference:}
Upper: three examples of Adenomatous polyps. (b) Lower: three  examples of Hyperplastic polyps. They are visually very similar although from different categories.}
\label{fig:class_examples}
\end{center}
\end{figure*}

Colonoscopy is the main diagnostic procedure to detect and recognize polyps located on colorectal walls. The accurate detection and correct classification depend on the skills and experience of the endoscopists, however, even for experienced endoscopists, working on conventional colonoscopy for long hours leads to mental and physical fatigue and degraded analysis and diagnosis. Other factors that may affect the classification results include varying illumination conditions, variant texture and appearance, and occlusion. Moreover, different types of polyps are hard to differentiate since they may exhibit a very similar appearance with a subtle difference, as shown in Fig~\ref{fig:class_examples}. It requires a thorough examination of fine details to distinguish one category form the other. Therefore, an accurate and effective automatic computer-aided system for colonoscopy is required to help endoscopists to detect and classify the type of polyps. This automated recognition mechanism can also be used as a second opinion to determine whether a further biopsy is required for diagnosis, which in turn will greatly reduce the cost of diagnosis. In addition, such an intelligent system can also be used as an educational resource for gastroenterology trainees to reduce the learning curve and cost. 

In recent years, deep learning algorithms have shown their outstanding performance on various generic datasets \cite{li2019object}. In some computer vision tasks, including strategic board games, Atari games, and generic object recognition, deep learning even outperforms human accuracy. However, there is a significant difference between generic images and medical images, as medical images contain more quantitative information and the object have no canonical orientation. In addition, acquiring medical data is expensive and labeling them requires the involvement of domain experts. In this work, although we have used a total of 27,048 images to train our models, they are extracted from only 119 video sequences with each sequence contains one polyp. In short, we have only 119 different polyp images taken from various viewpoints with varying lighting conditions to train our models.

Based on the result of our previous study \cite{mo2018efficient}\cite{li2019polyp} and the results of MICCAI Endoscopic Vision Challenge \cite{bernal2017comparative}, we can see that the state-of-the-art object detection models can already yield a very high precision in polyp detection. In this study, we assume the polyps have been detected and focus our study only on classification. 

\begin{figure*}[!h]
\begin{center}
\begin{tabular}{ccc}
 \includegraphics[width=38mm,height=32mm]{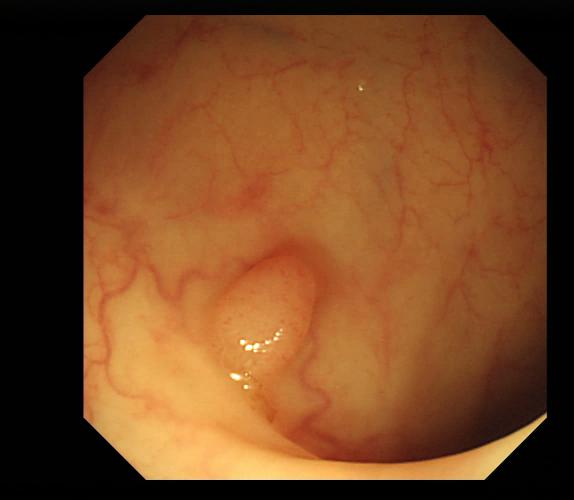} &   \includegraphics[width=38mm,height=32mm]{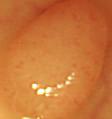}  &  \includegraphics[width=38mm,height=32mm]{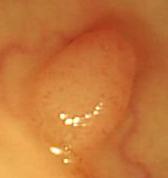}  \\
(a)& (b) & (c)  \\ [6pt]

\end{tabular}
\caption{{\bf Type of polyp input: }Same polyp frame with different versions of input. (a) Full frame, where the actual polyp feature is less compare to its background features. (b) The cropped polyp. (c) The cropped polyp with around 55\% of background. We generate data set-1 using (b) and set-2 using (c) in this study.}
\label{fig:typeofinput}
\end{center}
\end{figure*}

In our previous work \cite{li2019polyp}, we have collected and annotated a collection of endoscopic dataset, which contains 157 video sequences and a total of 35,981 frames. We have also labeled the ground truth of the polyp location and histogram class. In order to evaluate the performance of different classification models, we generate two polyp datasets from the annotated endoscopic dataset. As shown in Fig~\ref{fig:typeofinput}, one dataset (set-1) only contains the cropped polyp patches from the original video frames; the other dataset (set-2) contains not only the cropped polyps but also around 55\% background around the polyps. As described in \cite{nice}, polyps have different surrounding and vascular patterns and color in vessels and background according to the type of polyps. Therefore, we generate set-2 to study the effect of background features \cite{nice} in polyp classification. 

Fig~\ref{fig:typeofinput} illustrates the difference between the two generated datasets. We have evaluated and compared the performance of six classification models on these two datasets. Our results show that there is no significant difference in classification accuracy between the two datasets. We have also analyzed the performance based on both individual frames and individual sequences. The major contribution of this work include:

\begin{itemize} 
\item We have generated two datasets for polyp classification. To the best of our knowledge, there are no such datasets available in the literature,
\item we have implemented six state-of-the-art deep learning-based image classification models and compared their performance on the two datasets. This is the first comparative evaluation for polyp classification using different convolutional neural network (CNN) models.
\item This study can serve as a baseline for future studies on polyp classification. The trained classification models, as well as the test dataset will be available for free to the research community on the author's website. 
\end{itemize}

\subsection*{Related Work}
Various approaches and models have been proposed for polyp detection in colonoscopy. Previous comparative validation study on MICCAI 2015 polyp detection challenge shows the proposed models using handcrafted features as well as deep learning models. However, to the best of our knowledge, most previous works were focused on polyp detection, rather than classification, due to the unavailability of the dataset. There have been very few models proposed for polyp classification which classify the polyp into the hyperplastic and adenomatous type. Previous polyp classification approaches can be broadly divided into two categories: handcrafted feature based and deep learning based model.

\textbf{Conventional Computer Vision Approaches:} Most of the polyp classification work in the literature are based on handcrafted features.  Some approaches employ a pit pattern classification scheme to classify the polyp \cite{wimmer2016evaluation} into two classes: normal mucosa and hyperplastic. Hafner {\it et al.} \cite{hafner2015local} went beyond the conventional pit patterns approach and exploited fractal dimension based (LFD) strategy. Uhl {\it et al.} proposed a blob-adapted local fractal dimension(BA-LFD) approach \cite{uhl2014shape} to classifying polyps. Maximal-minimal filter bank strategy proposed by \cite{wimmer2016novel} outperformed the BA-LFD based approach. 

\textbf{Neural Network Based Approaches:} The study \cite{ribeiro2016exploring} provided a first review of various deep learning based models for polyp classification. They compared the performance of VGG-VD \cite{simonyan2014very}, CNN-F \cite{chatfield2014return}, CNN-M \cite{chatfield2014return}, CNN-S \cite{chatfield2014return}, AlexNet \cite{krizhevsky2012imagenet}, and GoogleLeNet \cite{szegedy2015going} on i-Scan1, i-Scan2 and i-Scan3 database. The paper \cite{korbar2017deep} utilized CNN model to classify the polyp, but in their experiments they employed whole side images instead. The study \cite{akbari2018classification} classified the polyps into informative and non informative categories instead of hyperplastic and adenomatous. 

\textbf{Deep Learning Models:}
Inspired by the success of AlexNet \cite{krizhevsky2012imagenet} in the ImageNet Large Scale Visual Recognition Challenge (ILSVRC) 2012, convolutional neural networks (CNN) have attracted a lot of attention and been successfully applied to image classification \cite{cen2019boosting}\cite{cen2019dictionary}\cite{wu2019unsupervised}, object detection \cite{ma2020mdfn}\cite{li2019object}\cite{ma2018mdcn}, depth estimation \cite{he2018learning}\cite{he2018spindle}, image transformation \cite{xu2019toward}\cite{xu2019adversarially}, and crowd counting \cite{sajid2020zoomcount}cite{sajid2020plug}. VGGNets \cite{simonyan2014very}, and GoogleNet \cite{szegedy2015going}, the ILSVRC winners of 2014 and 2015, proved that deeper models could significantly increase the ability of representations. ResNet \cite{he2016deep} proposed a skip connection based residual module to solve the vanishing gradient problem of very deep models. Highway networks \cite{srivastava2015highway} proposed a gating mechanism to regulate the flow of information in short - connections. ResNetxt \cite{xie2017aggregated} proposed to employ multi-branch architecture and proved the cardinality as an essential factor in the CNN architecture. Huang {\it et al.} proposed DenseNet \cite{huang2017densely} where each layer is connected to all subsequent layers. The winner of ILSVRC 2017, SENet \cite{hu2018squeeze}, achieved 82.7\% top-1 accuracy by improving channel interdependencies at almost no computational cost. Recently, EfficientNet \cite{tan2019efficientnet} has been proposed, which introduced a new scaling method for CNN and achieved improved performance.

Most of the proposed CNN models are based on the following three approaches: (1) Increasing the depth (number of layers) and/or width of the block architecture; (2) introducing an attention module; and (3) using a neural architecture search mechanism. 
The models chosen in this work are the classical models using all these three approaches. In the task of object detection, classification models are used as a backbone network, and the performance of object detection largely relied on the backbone network. The most widely adopted backbone networks including VGG, ResNet, and DenseNet. Therefore, we include all these three models in our study. In addition, we also include SENet and MnasNet. SENet employs a novel channel-wise attention mechanism, while MnasNet uses a neural architecture search. These models will demonstrate the performance of the state-of-the-art CNN models in polyp classification.


\section*{Materials and methods}

Convolutional neural networks have been widely applied to various computer vision tasks including object detection and classification. A general CNN network consists of different blocks, including an input layer, an output layer, and a number of hidden layers made up of convolution layers, pooling layers, and activation layers. CNNs adaptively learn spatial hierarchies of features via back propagation through these building blocks. In this section, we make a brief review of the classical object classification models used in this comparative study. These models include VGG \cite{simonyan2014very}, ResNet \cite{he2016deep}, DenseNet \cite{huang2017densely},  Squeeze-and-Excitation Network (SENet) \cite{hu2018squeeze} and MnasNet \cite{tan2019mnasnet}. 

\subsubsection*{VGG}
VGG Net \cite{simonyan2014very} was proposed by Simonyan and Zisserman to improve the classification performance by adding more convolutional layers to increase the depth of the network. This
could be possible by replacing a large filter size ($11 \times 11$ and 
$5 \times 5$) with  $3 \times 3$ multiple kernel sized filter stacked together. Max pooling layer is used to reduce spatial dimensions at every few layers. There are three back-to-back fully connected and a softmax layer respectively followed by stacking the $3 \times 3 $ convolution layers at the end. VGG is the first network structure that adopts block-based architecture. ReLU non-linearity has been added to all hidden layers. The number of weight parameters in VGG is larger than the previously proposed AlexNet, though it takes fewer epochs to converge because of implicit regularization imposed by its depth and small convolution filter size.
 
\subsubsection*{ResNet}

To address the problem of vanishing gradients in deep neural networks, He {\it at al.} \cite{he2016deep} proposed ResNet which was implemented using the idea of Residual - Blocks, with skip connection to fit the input from the previous layer to the next layer without modifying it. In addition, the residual block structure was structured for different deep variants of ResNet, ResNet-50, and ResNet-101, by including bottleneck design. For each residual block, they used a stack of 3 layers instead of 2 layers, which includes $ 1 \times 1$ convolution layer back and forth of $3 \times 3$ layer. Here $ 1\times 1 $ layer is responsible for adjusting the dimensions. Though ResNet is deeper than the VGG net, it has fewer filters and lower complexity. ResNet-34 has 3.6 billion Flops which is only 18 \% of VGG-19.

\subsubsection*{DenseNet}
Huang {\it at al.} \cite{huang2017densely}  proposed DenseNet based on the observation that deep network is efficient to train if they contain shorter connections between layers close to the input and layers close to the output. DenseNet is made up of several dense blocks and the feature maps from all previous layers are used as an input, and its own feature map is used as input to all subsequent layers. DenseNet uses concatenation operation to add the features from previous layers instead of using element-wise addition. In DenseNet, each layer has fewer number of filters(12 filters), which makes the network thinner and compact. In addition to fewer weight parameters, DenseNet is easy to train because of improved information flow and gradients throughout the network.

As each layer produces $k$ feature maps. $1 \times 1$ convolution layer is used to reduce the number of input feature map before applying it to a $3\times 3 $ convolution layer. With this unique design architecture, DenseNet has succeeded to reduce the vanishing gradient problem as well as strengthen feature propagation and encourage feature reuse. 

\subsubsection*{SENet}
Researchers have tried to improve the accuracy by stacking layers in different ways. Hu {\it at al.} \cite{hu2018squeeze} proposed a new architecture block squeeze and excitation based on the observation that not all feature maps are equally important. In conventional convolutional networks, the output feature maps are equally weighted, whereas SENet block weights each channel adaptively in a kind of content-aware mechanism. In more formal terms: SE block employs global information to selectively emphasize informative features and suppress less useful ones. The SE block is made up of two different operations: Squeeze and excitation. The squeeze operation uses global average pooling to generate channel-wise statistics which is a $n$-dimensional feature vector where $n$ is the number of channels. The excitation operation utilizes this $n$-dimensional feature vector, passes through two fully connected layers, and generates the same length vector. This resultant vector is used to weight the original feature maps. This squeeze and excitation block can be embedded into any state-of-the-art object classification models at a slightly additional cost. The squeeze and excitation network won the first place in ILSVRC 2017 classification and reduced the top-5 error to 2.251\%. 

\subsubsection*{MnasNet}
MnasNet \cite{tan2019mnasnet}, proposed by Google Brain, is an automated mobile neural architecture search approach, based on reinforcement learning, which can identify a model that could achieve a good trade-off between accuracy and latency. MnasNet introduced  a hierarchical  search space that provides layer diversity throughout the network instead of repeatedly stack the same cells through the network. The main components of MnasNet include (i) RNN-Controller  used for sampling model architecture; (ii) a trainer used to trained model sampled by RNN-controller; and (iii) a mobile phone-based inference engine for measuring latency.  MnasNet has been implemented on the ImageNet \cite{deng2009imagenet} and COCO \cite{lin2014microsoft} database. In this work, we used the architecture which was searched by MnasNet on the ImageNet\cite{deng2009imagenet} dataset.

\subsection*{Implementation}
\subsubsection*{Dataset Preparation}
In order to evaluate the performance of different models on the classification of polyps. We collected and labelled the following datasets.
\begin{enumerate}

\item \textbf{MICCAI 2017 Dataset:}
This dataset was published at the GIANA Endoscopic Vision Challenge held at MICCAI 2017. It contains 18 short videos for training and 20 videos for testing\cite{bernal2017comparative}. Each frame in the training set has its associated ground truth in the form of segmentation mask.
\item \textbf{CVC ColonDB Dataset:}
This dataset was published by Bernel {\it at al.} \cite{bernal2012towards}, which contains 15 short colonoscopy video sequence, with the ground truth of polyp segmentation mask. 
\item \textbf{ISIT-UMR Colonoscopy Dataset:}
This dataset was published by Mesejo {\it at al.} \cite{mesejo2016computer}. It contains 76 short video sequences. Each video sequence was labeled by the polyp categories, however, there is no ground truth of segmentation. 
\item \textbf{KUMC Colonoscopy Dataset:}
This is a dataset collected at the University of Kansas Medical Center with ethical oversight . It consist of 80 colonoscopy video sequences. 
\end{enumerate}

With the help of three endoscopists from the medical school of Jilin University and the University of Kansas Medical Center, we labeled the polyp classification results of all videos in datasets 1, 2, and 4. We also annotated the location bounding boxes for all the polyps in datasets 3 and 4. During the annotation process, the endoscopists could not reach an agreement on some sequences since they may need further biopsy verification. Those videos are removed from the datasets. We finally obtained a dataset of 157 videos (35,981 frames) with the labeled ground truth of the polyp histology and bounding boxes. 

For the labeled dataset, we randomly split all the videos into training, validation, and test sets which contains 119, 16, and 22 video sequences, respectively. 
The study focuses on evaluating the performance of the state-of-the-art classification models. We assume the polyps have been accurately detected and generate two separate datasets for the evaluation. As shown in  Fig~\ref{fig:typeofinput}, set-1 only contains the patches of the cropped polyps, and set-2 contains not only the cropped polyps but also about 55\% background around the polyps.

\subsection*{Training}

In this study, we implemented and compared a total of 6 classical classification models:  VGG19 with/without batch normalization \cite{simonyan2014very}, ResNet50 \cite{he2016deep}, DenseNet121 \cite{huang2017densely}, SE-ResNet50 \cite{hu2018squeeze} and MnasNet \cite{tan2019mnasnet}. The training dataset contains 119 sequences (27,048 images). 
We train all the model using NVIDIA Tesla K80 or P100 GPUs. The hyperparameters used to train the models are tabulated in Table \ref{tab:3}. All models were initialized by pre-trained ImageNet weights and the training time of each model ranges from 1 to 3 hours.

\begin{table*}[!ht]
\centering
\caption{{\bf Hyperparameters}}
\begin{tabular}{|l+l|l|l|l|l|}
 \hline
Model&Learning rate& Batch size&Epoch&Step size&Gamma\\
\thickhline
VGG19 & 0.001&32&25&-&-\\
VGG19-BN&0.001&32&25&-&-\\
ResNet50&0.001&64&25&-&-\\
DenseNet&0.001&64&25&-&-\\
SE-ResNet&0.001&64&50&30&0.1\\
MnasNet&0.001&64&150&-&-\\
\hline
\end{tabular}
\begin{flushleft} The hyperparameters used to train different models.
\end{flushleft}
\label{tab:3}
\end{table*}

\subsection*{Evaluation Metrics}
In the experiments, we train each model until it achieves the optimal performance on the validation set. To evaluate the model performance, we calculate the top-1 classification error. In order to make a fair comparison of different models, the performance has also been evaluated in terms of sensitivity, specificity, accuracy, precision, and F1-Score. The definitions of these matrices are listed in Table \ref{tab:def}. 
We evaluates the performance of all models on each sequences individually for both datasets.

\begin{table*}[!ht]
\begin{adjustwidth}{-2.25in}{0in}
\centering
\caption{{\bf Evaluation Metrics}}
\begin{tabular}{|l+p{12cm}|}
\hline
 & Polyp classification \\
\thickhline
True Positive(TP) & Numbers of adenomatous polyps that are correctly classified \\
True Negative(TN)& Numbers of  hyperplastic polyps that are correctly classified\\
False Positive(FP)& Numbers of hyperplastic polyps that are incorrectly misclassified as adenomatous\\
False Negative(FN) & Numbers of adenomatous polyps that are incorrectly classified as hyperplastic\\
Sensitivity&\% of actual adenoma are correctly classified. Also termed as recall and accuracy of adenoma. $\frac{TP}{TP + FN} \times 100$\\
Specificity& \% of actual hyperplastic are correctly classified. Also termed as recall and accuracy of hyperplastic. $\frac{TN}{TN + FP} \times 100$\\
Precision(Adenoma)&\% of predicted adenoma that are truly adenoma. $\frac{TP}{TP + FP} \times 100 $\\
Precision(Hyperplastic)&\% of predicted hyperplastic that are truly hyperplastic. $\frac{TP}{TP + FP} \times 100 $\\
Accuracy&Overall accuracy of both classes. $\frac{TP + TN}{TP + TN + FP + FN}\times 100$\\
F1-Score& Weighted average of precision and recall. $2\frac{precision \times recall}{precision + recall }\times 100$\\
Error&$\frac{1 - Accuracy} { 100}$\\
ROC & Receiver operating characteristic curve \\
AUC &Area under the curve (of ROC)\\
\hline
\end{tabular}
\begin{flushleft} Evaluation metrics used in the comparison. Precision, Recall(class based accuracy) and  F1-Score are calculated for both classes
\end{flushleft}
\label{tab:def}
\end{adjustwidth}
\end{table*}

\section*{Results}
In this section, we report the classification results of all comparative models using the two datasets. All input images are resized to $224 \times 224$ for a fair comparison. All models include batch normalization except VGG-19. The test set contains a total of 22 sequences (4719 frames), where 13 sequences (2890 frames) belong to adenomatous and 9 sequences (1829 frames) belong to hyperplastic. All models employ softmax as the classifier to yield the scores for the two classes, and the model outputs the class corresponding to the higher score. The top-1 error, precision, recall (individual class accuracy), and F1-score for both categories are as shown in Table  \ref{tab:confmat}. To alleviate the influence of the variation of illumination, all images in the datasets were normalized with respect to their mean and standard deviation. The mean and standard deviation of both datasets are listed in Table \ref{tab:4}.

\begin{table*}[!ht]
\begin{adjustwidth}{-2.25in}{0in}
\centering
\caption{{\bf Evaluation Results}}

\begin{tabular}{|l+p{0.4cm}p{0.4cm}p{0.4cm}p{0.4cm}p{0.8cm}p{1.1cm}p{0.9cm}p{1.2cm}p{1.2cm}p{1.2cm}p{1.2cm}p{1.2cm}p{0.8cm}|}
\hline
Model&TP&TN&FP&FN&Ade&Hyper&Acc&Err&Pre-1&Pre-2&F1-1&F1-2&AUC\\
 &&&&&(\%)&(\%)&(\%)&(\%)&(\%)&(\%)&(\%)&(\%)&(\%)\\   
\thickhline
VGG-19(set-1)&2424&1149&680&466&\textbf{83.87}&62.82&\textbf{75.71}&\textbf{24.28}&78.09&\textbf{71.14}&\textbf{80.88}&66.72&76.43\\
VGG-19(set-2)&2419&1346&483&471&\textbf{83.70}&\textbf{73.59}&\textbf{79.78}&\textbf{20.21}&\textbf{83.35}&\textbf{74.07}&\textbf{83.52}&\textbf{73.83}&84.80\\
\hline
VGG19-BN(set-1)&2071&1440&389&819&71.66&\textbf{78.73}&74.40&25.59&\textbf{84.18}&63.74&77.42&\textbf{70.45}&78.58\\
VGG19-BN(set-2)&2295&1345&484&595&79.41&73.53&77.13&22.86&82.58&69.32&80.96&71.37&82.20\\
\hline
ResNet50(set-1)&2350&1222&607&540&81.31&66.81&75.69&24.30&79.47&69.35&80.38&68.05&77.25\\
ResNet50(set-2)&2042&1305&524&848&70.65&71.35&70.92&29.07&79.57&60.61&74.85&65.54&76.27\\
\hline
DenseNet(set-1)&2246&1282&547&644&77.71&70.09&74.76&25.23&80.41&66.56&79.042&68.28&79.28\\
DenseNet(set-2)&2065&1306&523&825&71.45&71.40&71.43&28.56&79.79&61.28&75.39&65.95&78.65\\
\hline
SENet(set-1)&2230&1320&509&660&77.16&72.17&75.22&24.77&81.41&66.66&79.23&69.30&72.78\\
SENet(set-2)&2338&1138&691&552&80.89&62.21&73.65&26.34&77.18&62.21&78.99&64.67&82.05\\
\hline
MnasNet(set-1)&2239&1213&616&651&77.47&66.32&73.15&26.84&78.42&65.07&77.94&65.69&73.32\\
MnasNet(set-2)&2115&1242&587&775&73.18&67.90&71.13&28.86&78.27&61.57&75.64&64.58&77.11\\
\hline
\end{tabular}

\begin{flushleft}
Overall performance of all model on set-1 and set-2 based on individual frame irrespective of sequence.
\end{flushleft}
\label{tab:confmat}
\end{adjustwidth}
\end{table*}

\begin{table*}[!ht]
\centering
\caption{{\bf Mean and standard deviation}}
\begin{tabular}{|l+p{7.3cm}|}
\hline
  & Mean and standard deviation used for normalization\\
\thickhline
Set-1 & [0.6916, 0.5297, 0.4158][0.1439, 0.1377, 0.1306] \\
Set-2 & [0.6594, 0.5112, 0.4026][0.2469,0.2254,0.2095]\\
\hline
\end{tabular}
\begin{flushleft} Mean and standard deviation of set-1 and set-2, used to normalize input images.
\label{tab:4}
\end{flushleft}
\end{table*}

\section*{Discussion}

\subsection*{Frame-based Performance}
We first report the comparative performance of different models based on each individual frame. Frame-based performance is measured without considering the particular sequence of those frames. It measures the overall accuracy similar to the generic classification evaluation for other datasets.
As shown in Table \ref{tab:confmat}, VGG19 outperforms all other models with an overall accuracy of 75.71\% and 79.78\% for set-1 and set-2, respectively. The precision of Adenomatous class is higher than that of Hyperplastic class for every model in both datasets, except for VGG-19 with batch normalization (on set-1) and ResNet50 (on set-2).  If we consider precision and F1-score for every model in both datasets, the precision of Adenomatous is always higher than that of Hyperplastic. VGG-19 has also achieved the highest recall for both classes on set-2. The most recently proposed models, like ResNet, SENet, and MnasNet did not perform well in both datasets, although they have better performance than VGG-19 on generic image classification datasets. 

From Table \ref{tab:confmat} we also observe that VGG-19 outperforms VGG-19 with batch normalization in most metrics. This is contradicting to what was observed in other datasets. The reason might because that, in polyp classification, the exact intensity values of the pixels may be more useful for the discrimination of different types of polyps than that in generic image classification. While batch normalization layer scales the pixel values with respect to the batches, which may affect the intensity information and downgrade the performance. 

\begin{figure*}[!h]
\begin{adjustwidth}{-2.25in}{0in}
\begin{center}
\begin{tabular}{ccc}
\includegraphics[width=55mm,height=55mm]{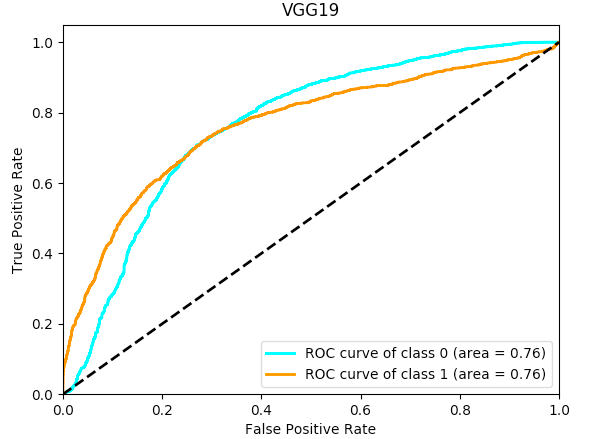} &\includegraphics[width=55mm,height=55mm]{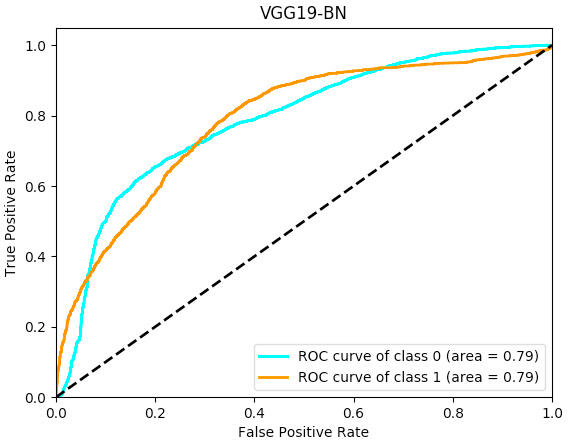} & \includegraphics[width=55mm,height=55mm]{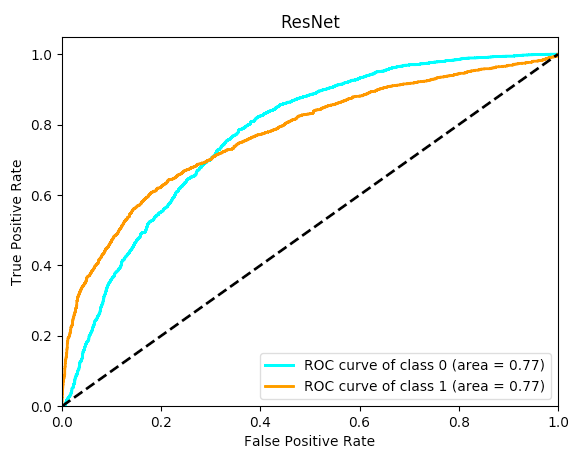} \\
(a)&(b)&(c) \\
\includegraphics[width=55mm,height=55mm]{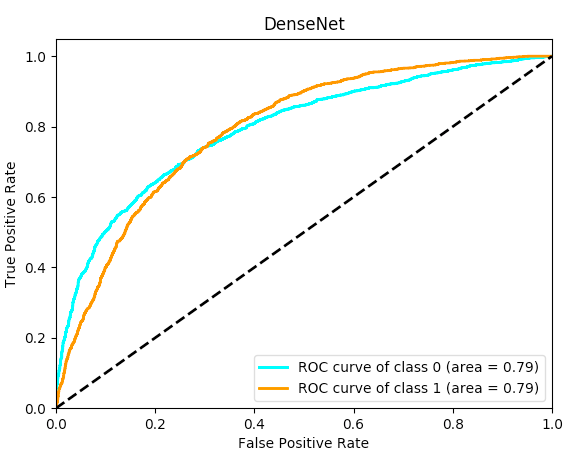} &\includegraphics[width=55mm,height=55mm]{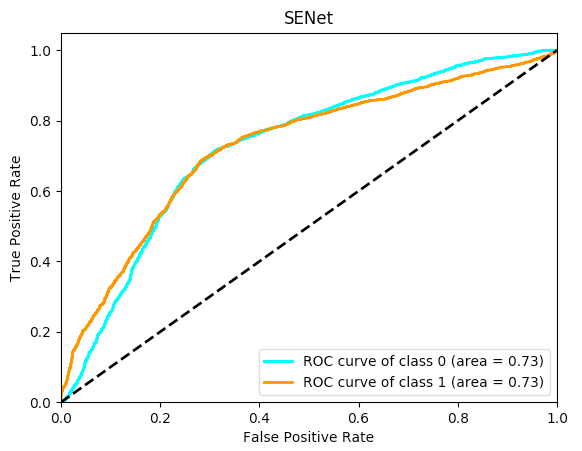} &\includegraphics[width=55mm,height=55mm]{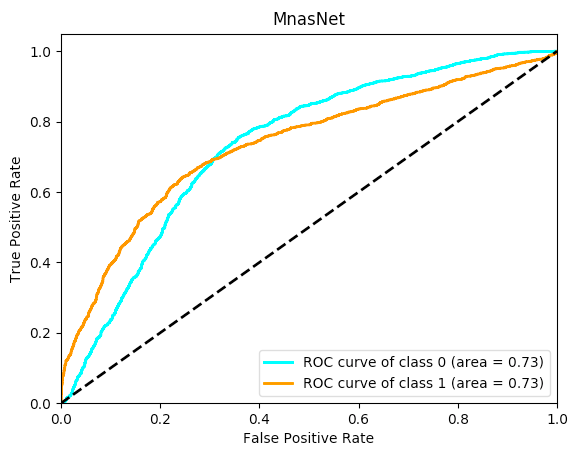} \\
(d)&(e)&(f) \\
\end{tabular}
\end{center}
\caption{{\bf AUC-ROC curves of different models on set-1:} (a) VGG19 (b) VGG19-BN (c) ResNet50 (d) DenseNet (e) SENet (f) MnasNet}
\label{fig:Roccrop}
\end{adjustwidth}
\end{figure*}

\begin{figure*}[!h]  
\begin{adjustwidth}{-2.25in}{0in}
\begin{center}
\begin{tabular}{ccc}
 \includegraphics[width=55mm,height=55mm]{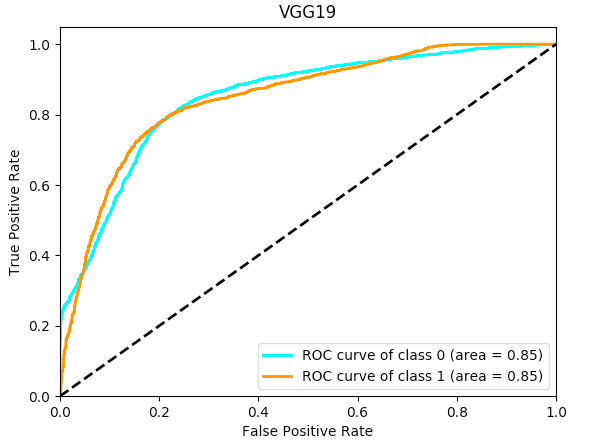} &\includegraphics[width=55mm,height=55mm]{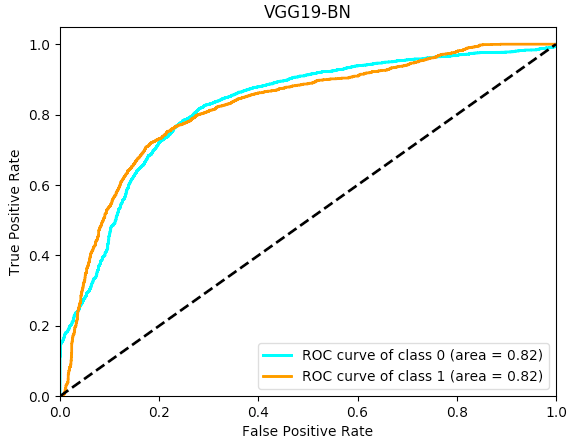} & \includegraphics[width=55mm,height=55mm]{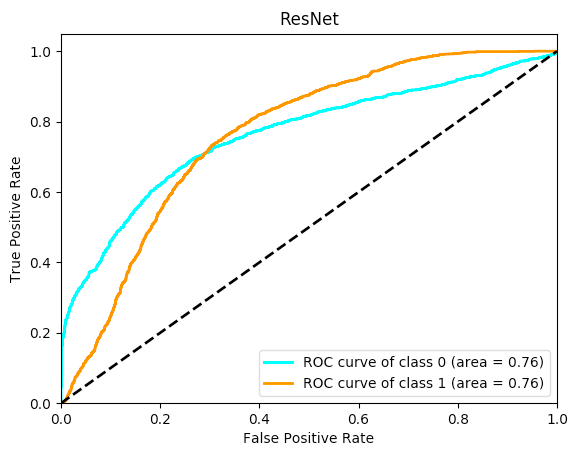} \\
(a)&(b)&(c) \\
\includegraphics[width=55mm,height=55mm]{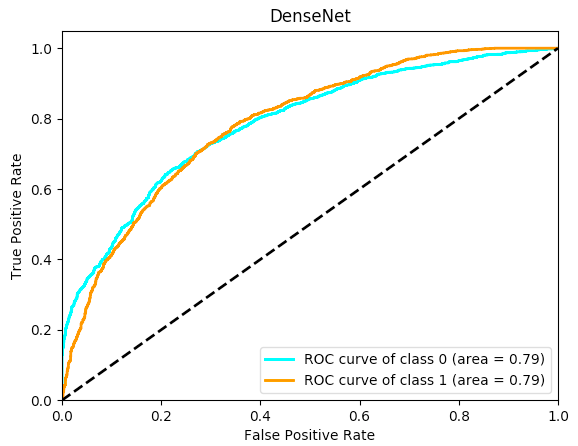} &\includegraphics[width=55mm,height=55mm]{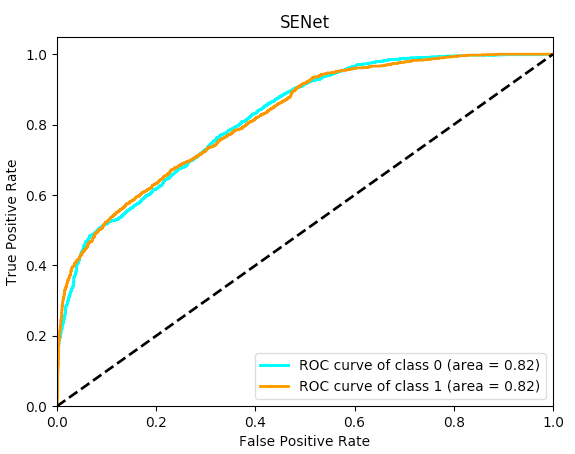} &\includegraphics[width=55mm,height=55mm]{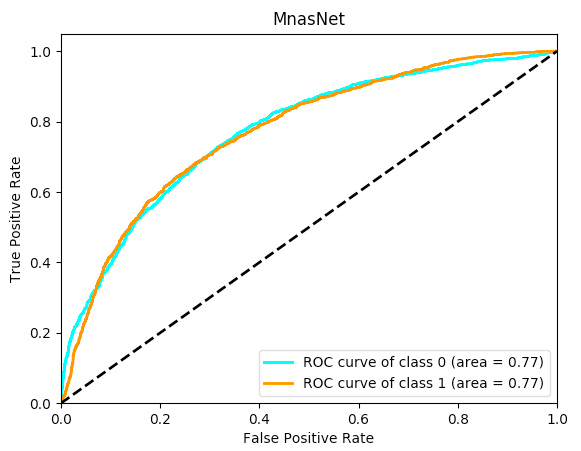} \\
(d)&(e)&(f) \\
\end{tabular}
\end{center}
\caption{{\bf AUC-ROC curves of different models on set-2:} (a) VGG19 (b) VGG19-BN (c) ResNet50 (d) DenseNet (e) SENet (f) MnasNet}
\label{fig:rocback}
\end{adjustwidth}
\end{figure*}

To better visulize the performance, we employ AUC (area under the curve) ROC (receiver operating characteristics) curve to demonstrate the frame-based performance. AUC-ROC curve represents the degree of separability of a classification problem. It demonstrates the capability of a model in differentiating classes. Fig~ \ref{fig:Roccrop} and Fig~ \ref{fig:rocback} show the ROC curves of different models for set-1 and set-2, respectively. The results show that, in general, the models achieve better classification performance on set-2 than that on set-1 except for ResNet. We can also see that VGG-19 achieves the highest ROC score and the best accuracy on set-2.

\subsection*{Sequence-based Performance}

\begin{figure*}[!ht]
\begin{adjustwidth}{-2.25in}{0in}
\begin{center}
\includegraphics[width=0.8\linewidth]{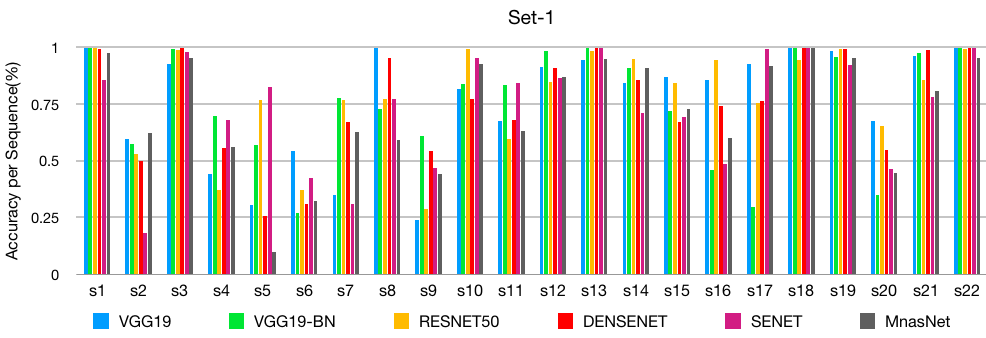}
\end{center}
   \caption{{\bf Sequence-based performance of set-1: }The performance of different models for each test sequence of set-1.}
\label{fig:cropseq}
\end{adjustwidth}
\end{figure*}

\begin{figure*}[!ht]
\begin{adjustwidth}{-2.25in}{0in}
\begin{center}
 \includegraphics[width=0.8\linewidth]{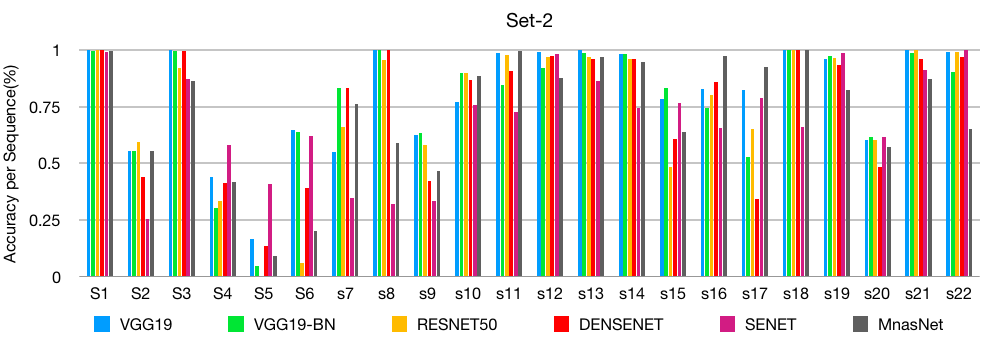}
 \caption{ {\bf Sequence-based performance of set-2: }The performance of different models for each test sequence of set-2.}
\label{fig:backseq}
\end{center}
\end{adjustwidth}
\end{figure*}

\begin{figure*}[!h]
\begin{adjustwidth}{-2.25in}{0in}
\begin{center}
\begin{tabular}{cccccc}
 \includegraphics[width=25mm,height=25mm]{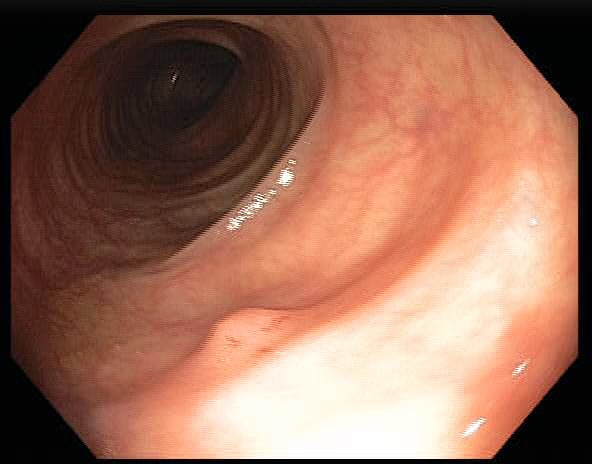} &\includegraphics[width=25mm,height=25mm]{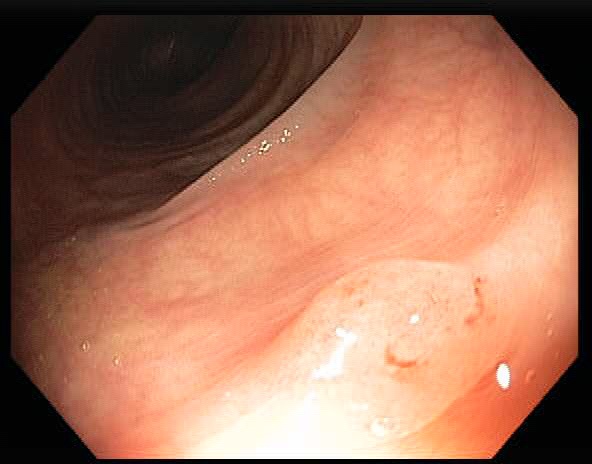} & \includegraphics[width=25mm,height=25mm]{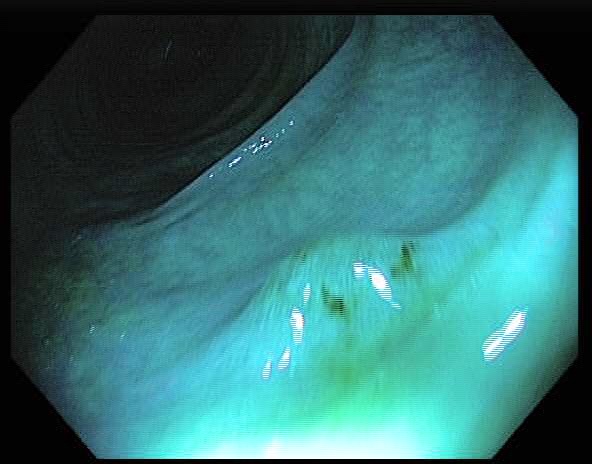}&
\includegraphics[width=25mm,height=25mm]{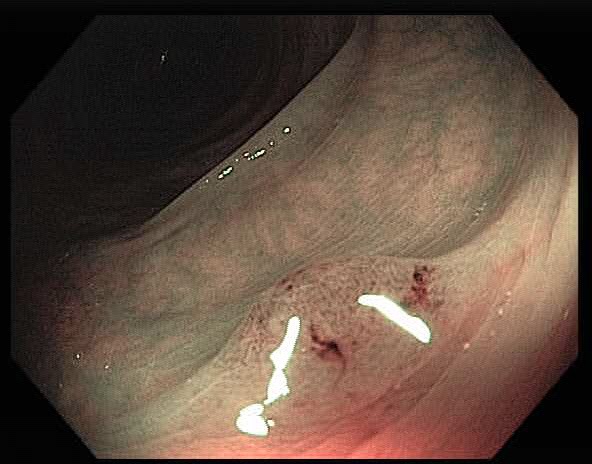} &\includegraphics[width=25mm,height=25mm]{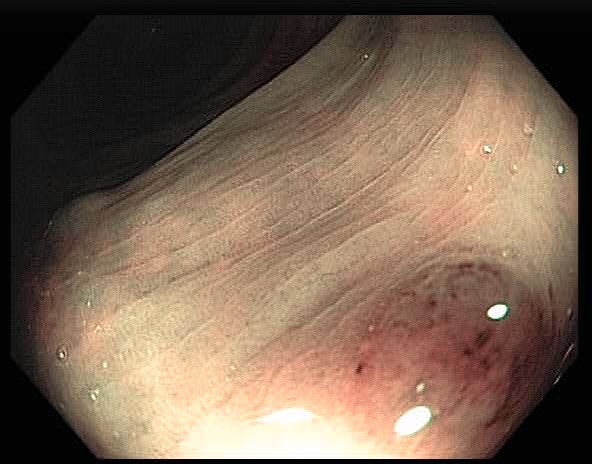} &\includegraphics[width=25mm,height=25mm]{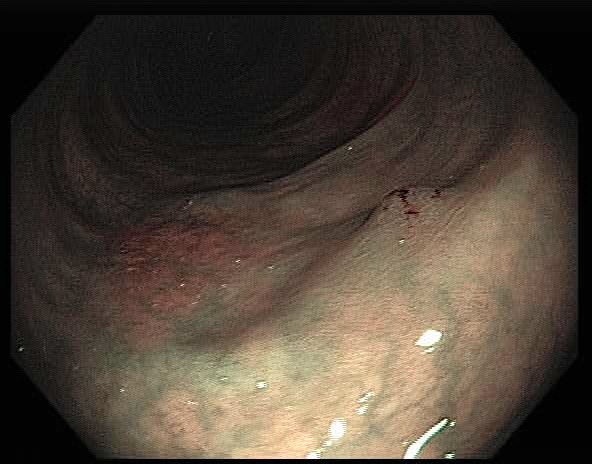} \\
(a)&(b)&(c)&(d)&(e)&(f)\\ [6pt]
\end{tabular}
\end{center}
\caption{{\bf Images from different viewpoints: }Six sample frames from the same sequence. The same polyp looks considerably different due to the variations of viewpoints and lighting conditions.}
\label{fig:viewpoint}
\end{adjustwidth}
\end{figure*}

Based on the classification of each frame, we can measure the performance of each sequence. The sequence-by-sequence performance for the two datasets are shown in Fig~ \ref{fig:cropseq} and Fig~ \ref{fig:backseq}, respectively. We can see that the results are not consistent among all frames within the same sequence of the same polyp. This is because the appearance of the polyp may subject to significant appearance changes due to the variance of the viewpoints, zooming scales, and illumination. Fig~ \ref{fig:viewpoint} shows some sample frames of a sequence under different viewpoints and lighting conditions. In this case, even experienced endoscopists cannot make an accurate prediction from a single frame. As a result, not all frames can be correctly classified. In practice, we calculate the percentage of correctly classified frames for each sequence. Then, we set a threshold in terms of the percentage, and a sequence is considered to be correctly classified if the percentage of correctly classified frames is greater than the specified threshold. Table \ref{tab:thresbased} shows the performance corresponding to different thresholds for the two datasets.
\begin{table*}[!ht]
\centering
\caption{{\bf Sequence-based accuracy}}
\begin{tabular}{|l+ccc|}
\hline
Model & Threshold(70\%) & Threshold(60\%) & Threshold(50\%) \\
\thickhline
VGG-19 &63.63/\textbf{68.18}&72.72/\textbf{81.81}&81.81/\textbf{90.90}\\
VGG19-BN&\textbf{69.63}/\textbf{68.18}&72.72/\textbf{81.81}&81.81/90.90\\
ResNet50&68.18/59.09&\textbf{77.27}/72.72&\textbf{86.36}/81.81\\
DenseNet&59.09/63.63&72.72/68.18&\textbf{86.36}/68.18\\
SE-ResNet&63.63/54.54&72.72/72.72&72.72/77.27\\
MnasNet&54.54/54.54&68.18/68.18&81.81/81.81\\

\hline

\end{tabular}
\begin{flushleft} Accuracy per sequence for  all models based on different threshold with set-1 / set-2 . First term before '/' specifies accuracy for set-1 and and term after '/' indicates accuracy for set-2.
\end{flushleft}
\label{tab:thresbased}
\end{table*}

\begin{figure*}[!ht]
\begin{center}
\begin{tabular}{ccc}
  \includegraphics[width=28mm,height=28mm]{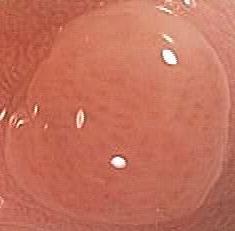} &\includegraphics[width=28mm,height=28mm]{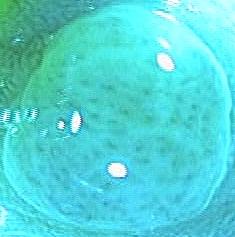} & \includegraphics[width=28mm,height=28mm]{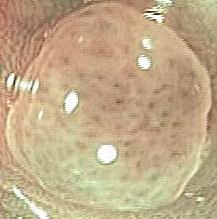}   \\
     &(a)& \\
 \includegraphics[width=28mm,height=28mm]{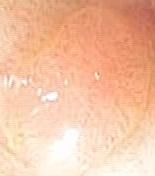} &\includegraphics[width=28mm,height=28mm]{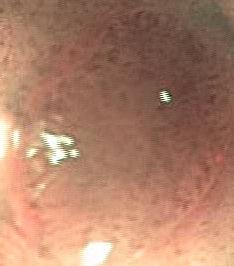} & \includegraphics[width=28mm,height=28mm]{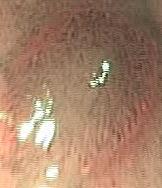}   \\
  &(b)&  \\ [6pt]

\end{tabular}
\caption{{\bf Missclassified sequences: }Sample frames from different sequences that could not be correctly classified by almost all models. (a) and (b) are sequences 5 and 6, respectively, where 5 is of type adenomatous and 6 is of type hyperplastic.}
\label{fig:exams}
\end{center}
\end{figure*}

As shown in Fig~ \ref{fig:cropseq} and Fig~ \ref{fig:backseq}, the classification result for each sequence is not consistent. The test sequences 1, 3, 10, 12, 13, 14, 18,19, 21, and 22 are correctly classified by all models for both datasets, while the results of sequences 2, 4, 5, 6, 7, 9, 11, 17, and 20 are not consistent because the percentage of the correctly classified frames is in between 40-50\%. Sequences  5 and 6 could not be classified well by all models. Some sample frames of sequences 5 and 6 are shown in Fig~\ref{fig:exams}, which subject large variations in appearance that cause the difficulty in classification.
Table \ref{tab:thresbased} shows the threshold-based performance of all models. The results indicate the consistency of the prediction of different models, from which we can see that VGG models achieve relatively better performance than other models. For example, VGG-19 achieves around 70\%, 80\%, and 90\% accuracy at the thresholds of 70\%, 60\%, and 50\%, respectively. Comparing Table \ref{tab:confmat} and Table \ref{tab:thresbased}, we can find that if we set the threshold at 50\%, the sequence-based accuracy is much higher than frame-based based accuracy, especially for VGG models. However, at a higher threshold of 70\%, the overall accuracy of the frame-based is higher than the sequence-based approaches, which indicates the consistent prediction within the sequence.

\begin{figure*}[!ht]
\begin{center}
\begin{adjustwidth}{-2.25in}{0in}
\begin{tabular}{cc}
 \includegraphics[width=0.5\linewidth]{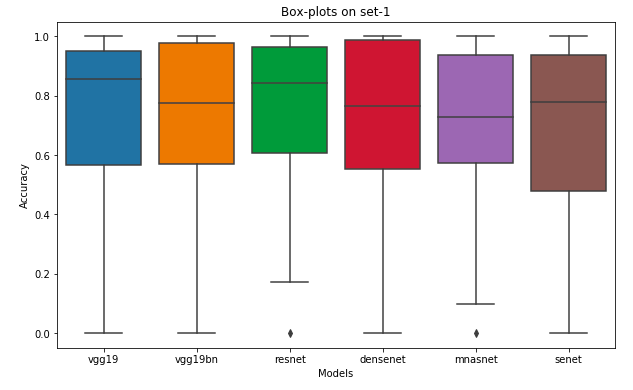}&
  \includegraphics[width=0.5\linewidth]{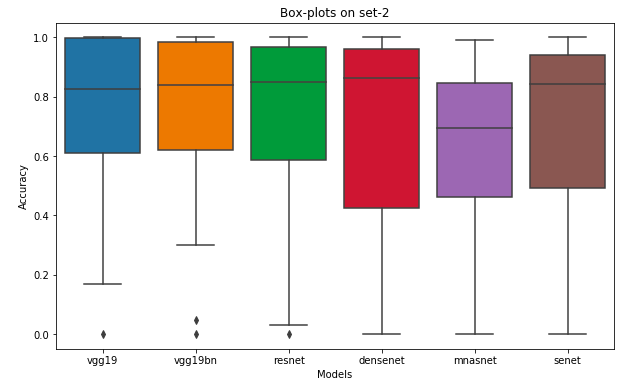} \\
  (a) & (b) \\
 \end{tabular}
\caption{{\bf Box plot of set-1 and set-2:} The accuracy per sequence distribution of different models on (a) set-1 and (b) set-2}
\label{fig:cropbox}
\end{adjustwidth}
\end{center}
\end{figure*}

To better visualize the sequence-based performance, we have included the box plots. Box plots show the accuracy per sequence distribution of the total 22 sequences. Fig~ \ref{fig:cropbox}  shows the box plots of different models on set-1 and set-2, respectively. It can be seen that the maximum accuracy of all models is 100\% because at least one sequence has been correctly classified by each of the models. The upper quartile range is dependent on the median value. A high median value decreases the upper half range, which shows the ability of the model to consistently correctly classified sequence. On set-1, VGG-19 achieves the highest median value, which indicates that half of the sequences are correctly classified with a very high threshold. On set-2, ResNet-50 yields the most consistent results with the highest median value. We can also see from the results that the upper quartile ranges are smaller than the lower quartile range, which indicates that the spread of accuracy below the median value is very high.

\subsection*{Polyp Crops vs Crops with Background}
In order to test the background information in polyp classification, we generate two datasets in the experiment, set-1 has only polyp crops and set-2 contains polyp crops with 50\% background. From Table \ref{tab:confmat} we can see that, if we consider frame-based performance, except for the VGG models, all other models achieve higher accuracy on set-1 than on set-2. If we consider the overall AUC-ROC score, set-2 yields better performance which means the two classes are easier to distinguish in set-2 than in set-1. If we consider sequence-based analysis, the performance of all sequences is almost similar for both types of datasets.  For consistency-based performance, the consistency is improved by VGG-19, VGG-19 with batch normalization, and DenseNet for set-2, whereas for other models, the overall threshold-based accuracy is very close. If we consider the box plots and set median as a threshold, the consistency of correctly classifying sequence is improved by ResNet, DenseNet, and SENet for set-2.

\section*{Conclusion}

In this paper, we have established two datasets and compared six state-of-the-art deep learning-based classification models. We have evaluated the results both at the frame level and at the polyp level. Our results show that VGG-19, in general, outperforms other models in both cases for both datasets. While some more advanced classification models, like ResNet, DenseNet, SENet, and MnasNet did not perform well in our experiments, though they have advantages on other benchmark datasets. The poor performance may be caused by the limited size of the polyp dataset. This study provides a good baseline for future research to develop more accurate and more robust polyp classification models.

\section*{Acknowledgement}
The authors would like to thank Dr. Vijay Kanakadandi at the University of Kansas Medical Center for his insightful help and advice for this study.

\nolinenumbers

\end{document}